\documentclass{article}

 \usepackage[dblblindworkshop, final]{neurips_2025}

\usepackage[utf8]{inputenc} 
\usepackage{fontenc}    
\usepackage{graphicx}    
\usepackage{amsmath}    
\usepackage{hyperref}       
\usepackage{algorithmic}       
\usepackage{algorithm}       
\usepackage{url}            
\usepackage{booktabs}       
\usepackage{amsfonts}       
\usepackage{nicefrac}       
\usepackage{microtype}      
\usepackage{xcolor}         
\usepackage{booktabs,multirow,makecell}
\usepackage[table]{xcolor}
\usepackage[table]{xcolor}
\usepackage{multirow}
\usepackage{booktabs}
\usepackage{graphicx}
\usepackage{wrapfig}    
\usepackage{natbib}

\definecolor{lightblue1}{HTML}{E0E8F8} 
\definecolor{lightblue2}{HTML}{C1D4F0}
\definecolor{lightblue3}{HTML}{A2C0E8}
\definecolor{lightblue4}{HTML}{83ACE0} 

\workshoptitle{Efficient Reasoning}

\title{M-GRPO: Stabilizing Self-Supervised Reinforcement Learning for Large Language Models with Momentum-Anchored Policy Optimization }

\author{Bizhe Bai$^{1,2}$, Hongming Wu $^{2}$, Peng Ye$^{3,4}$, Tao Chen$^{1,2}$, \\ \\
[1mm]
$^1$Shanghai Innovation Institute,
\\
$^2$College of Future Information Technology, Fudan.\\
$^3$ Shanghai AI Laboratory \\
$^4$ The Chinese University of Hong Kong \\
\texttt{\small {bizhe.bai@sii.edu.cn,eetchen@fudan.edu.cn}}
}

\begin{document}

\maketitle
\begin{abstract}
Self-supervised reinforcement learning (RL) presents a promising approach for enhancing the reasoning capabilities of Large Language Models (LLMs) without reliance on expensive human-annotated data. However, we find that existing methods suffer from a critical failure mode under long-horizon training: a "policy collapse" where performance precipitously degrades. We diagnose this instability and demonstrate that simply scaling the number of rollouts---a common strategy to improve performance---only delays, but does not prevent, this collapse. To counteract this instability, we first introduce M-GRPO (Momentum-Anchored Group Relative Policy Optimization), a framework that leverages a slowly evolving momentum model to provide a stable training target. In addition, we identify that this process is often accompanied by a rapid collapse in policy entropy, resulting in a prematurely confident and suboptimal policy. To specifically address this issue, we propose a second contribution: an adaptive filtering method based on the interquartile range (IQR) that dynamically prunes low-entropy trajectories, preserving essential policy diversity. Our extensive experiments on multiple reasoning benchmarks demonstrate that M-GRPO stabilizes the training process while the IQR filter prevents premature convergence. The combination of these two innovations leads to superior training stability and state-of-the-art performance. Code is available at \href{https://github.com/baibizhe/M_GRPO/tree/main}{M\_GRPO}.
\end{abstract}

\section{Introduction}

Reinforcement learning with   verifiable reward (RLVR) has become a central ingredient in post‑training large language models (LLMs) for complex reasoning tasks~\cite{shao2024deepseekmathpushinglimitsmathematical, yu2025dapoopensourcellmreinforcement}. While RLVR can substantially improve helpfulness and reliability, it is expensive and domain‑limited because it requires large amounts of carefully curated human preference data and reward modeling infrastructure. Recent work therefore explores self‑supervised or label‑free RL signals for reasoning: leveraging a model’s own uncertainty or self‑consistency to synthesize rewards and thereby train without ground‑truth answers or programmatic verifiers. They train on unlabeled prompts by constructing intrinsic signals from the model itself, such as self-consistency–derived correctness proxies or self‑certainty as a reward ~\cite{zhao2025learningreasonexternalrewards,zhang2025corewardselfsupervisedreinforcementlearning}. Some other methods include   test‑time reinforcement that turns majority‑vote statistics into a usable objective  such as SRT ~\cite{shafayat2025largereasoningmodelsselftrain} and TTRL ~\cite{Zuo2025TTRLTR}.

\begin{figure}{}{}    
\centering   
\includegraphics[width=0.95\textwidth, page=1]{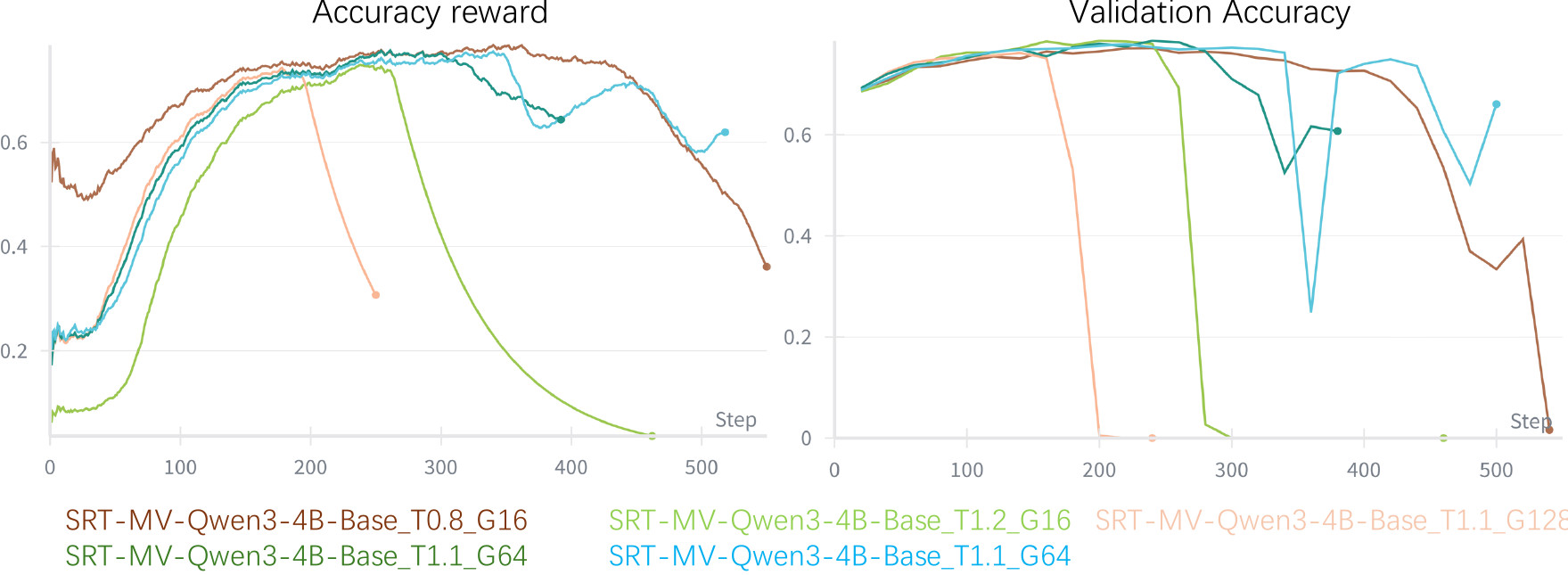}
\caption{
This caption details the reproduction of the SRT method ~\cite{shafayat2025largereasoningmodelsselftrain}. The experiments were conducted with different numbers of rollouts and a set temperature. For all methods, training was performed on the train split of the MATH dataset~\cite{hendrycks2021measuringmathematicalproblemsolving} without access to ground truth answers, using a consistent batch size and learning rate. The models were then validated on the corresponding MATH test split. Further test results, derived from the best-performing checkpoint chosen manually, are available in Table~\ref{tab:ssl_scaling_srt}} .
    \label{fig:srt-scaling-collapse}
\end{figure}

\paragraph{Self-supervised RLVR are easy to collapse}
These Self-supervised RLVR(SS-RLVR) approaches report promising early gains, but we find they suffer a critical failure mode under long-horizon training: the policy collapses.
In our reproduction of SRT~\cite{shafayat2025largereasoningmodelsselftrain}  and Intuitor~\cite{zhao2025learningreasonexternalrewards} self-supervised training on MATH dataset ~\cite{hendrycks2021measuringmathematicalproblemsolving} as shown in (Fig.~\ref{fig:srt-collapse}),  the training reward rises initially and then precipitously or progressively  crashes , accompanied by degradation in validation accuracy on the test set. This finding also aligns with the "model collapse" scenario with Sheikh's ~\cite{shafayat2025largereasoningmodelsselftrain} and Zizhuo's ~\cite{zhang2025corewardselfsupervisedreinforcementlearning}.
\vspace{-1em}
\paragraph{Scaling Rollouts Delays, But Does Not Prevent, Policy Collapse}
While increasing the number of rollouts during training is a known technique to bolster performance in supervised RLVR, as noted by recent work~\cite{tan2025scalingbehaviorsllmreinforcement,hu2025brorlscalingreinforcementlearning}, its application in the self-supervised context reveals a significant challenge. We investigate this scaling behavior and find a precarious trade-off: although more rollouts can improve the model's best performance (Table~\ref{tab:ssl_scaling_srt}), they only serve to slow the rate of descent into policy collapse, rather than preventing it entirely (Fig.~\ref{fig:srt-scaling-collapse}). This phenomenon suggests that while scaling the evidence pool for the self-rewarding mechanism can temporarily mitigate inherent instability, the model ultimately succumbs to the identified failure mode. This finding underscores the need for a stabilization method that can harness the benefits of larger rollout batches without eventually facing performance degradation.
\vspace{-1em}

\begin{table}[h]
\centering
\vspace{2mm}
\resizebox{\textwidth}{!}{
\begin{tabular}{l|ccccc}
\toprule[1.6pt]
\multirow{2}{*}{\textbf{}} & \multicolumn{3}{c}{\textbf{Math}} & \multicolumn{2}{c}{\textbf{Reasoning}}  \\
\cmidrule(lr){2-4}\cmidrule(lr){5-6}
~  & \textbf{MATH500 } & \textbf{AIME24 } & \textbf{AIME25 } & \textbf{GPQA Diamond} & \textbf{GPQA}  \\ 
\midrule
\multicolumn{6}{c}{\textit{\textbf{Qwen3-4B-base}}} \\
\midrule
Original & 0.615 & .00833 & 0.0500 & 0.3441 & 0.2991 \\
\midrule
SRT-T1.1-G16 & \cellcolor{lightblue1}{0.7515} & \cellcolor{lightblue1}{0.0917} & \cellcolor{lightblue2}{0.1042} & \cellcolor{lightblue1}{0.3706} & \cellcolor{lightblue1}{0.3482}  \\
SRT-T0.8-G16 & \cellcolor{lightblue1}{0.7395} & \cellcolor{lightblue1}{0.0958} & \cellcolor{lightblue1}{0.0708} & \cellcolor{lightblue2}{0.3655} & \cellcolor{lightblue4}{0.3772}  \\
SRT-T1.2-G16 & \cellcolor{lightblue2}{0.7615} & \cellcolor{lightblue3}{0.1042} & \cellcolor{lightblue1}{0.0708} & \cellcolor{lightblue3}{0.3649} & \cellcolor{lightblue2}{0.3571}  \\
SRT-T1.1-G32 & \cellcolor{lightblue1}{0.7405} & \cellcolor{lightblue3}{0.1083} & \cellcolor{lightblue3}{0.1062} & \cellcolor{lightblue1}{0.3586} & \cellcolor{lightblue1}{0.3348}  \\
SRT-T1.1-G64 & \cellcolor{lightblue4}{0.792} & \cellcolor{lightblue4}{0.1250} & \cellcolor{lightblue4}{0.1167} & \cellcolor{lightblue4}{0.3826} & \cellcolor{lightblue1}{0.3504} \\
SRT-T1.1-G128 & \cellcolor{lightblue3}{0.762} & \cellcolor{lightblue1}{0.0917} & \cellcolor{lightblue1}{0.0875} & \cellcolor{lightblue1}{0.3586} & \cellcolor{lightblue3}{0.3638}  \\
\bottomrule[1.6pt]
\end{tabular}}
\caption{Scaling result of rollout numbers G. Results with different rollout numbers and temperatures for SRT \cite{shafayat2025largereasoningmodelsselftrain}. 
    $T$ denotes the sampling temperature and $G$ the total number of rollouts (e.g., \texttt{SRT-T0.8-G16} means temperature $T=0.8$ and $G=16$ rollouts). 
    Although SRT improves at higher $G$, it remains prone to collapse as shown in Fig.~\ref{fig:srt-scaling-collapse}. All results are obtained by manually selecting the best checkpoint based on  accuracy reward before collapse.}
\label{tab:ssl_scaling_srt}
\end{table}

\begin{figure}[h]
    \centering
    \includegraphics[width=1\textwidth]{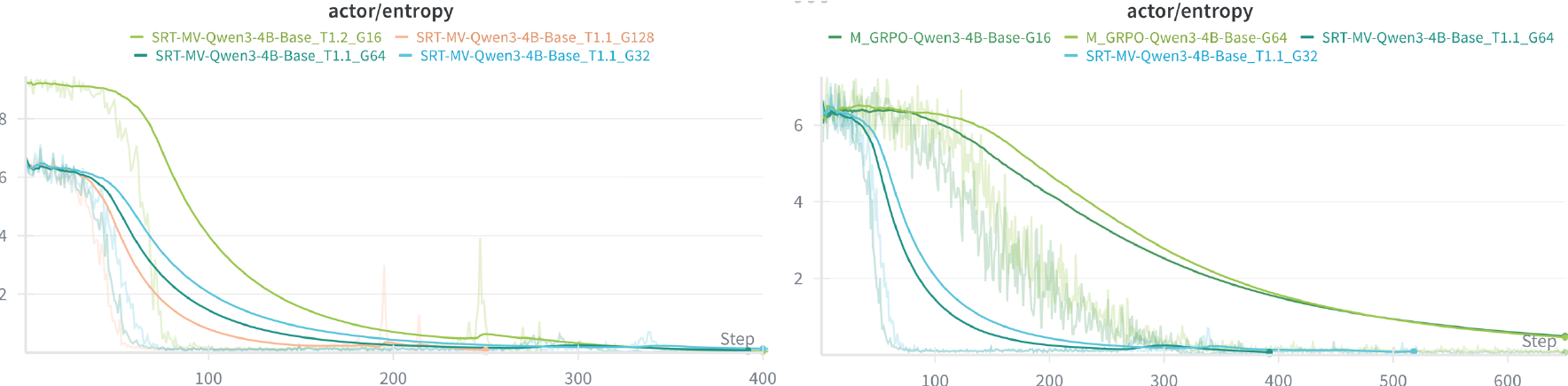}
    \caption{\textbf{Left:} The evolution of policy entropy during the training of a standard SRT model \cite{Zuo2025TTRLTR, shafayat2025largereasoningmodelsselftrain}. A sharp drop in entropy is observed early in training, leading to an overly confident policy. \textbf{Right:} A comparison with our proposed method, initialized  with M\_GRPO. Our approach maintains a higher level of entropy that decreases more gradually throughout the training process, mitigating premature convergence.}
    \label{fig:entropy}
\vspace{-1em}
\end{figure}

\paragraph{Self supervised reinforcement learning also caused entropy collapse} The phenomenon of policy entropy collapse \cite{cui2025entropymechanismreinforcementlearning} is also prevalent in self-supervised reinforcement learning, as illustrated on the left of Fig.~\ref{fig:entropy}. During the initial stages of training, the policy entropy declines precipitously, resulting in a prematurely confident policy model.

\paragraph{Contributions.}
First, we diagnose "policy collapse," a critical instability in self-supervised reinforcement learning for LLMs, and show that simply scaling rollouts exacerbates this failure. Second, we introduce M-GRPO, a novel momentum-anchored framework that leverages a slowly evolving model to provide a stable training target, effectively mitigating said collapse. Third, to prevent policy entropy collapse, we propose a filtering method based on interquartile range that adaptively filters the low-entropy trajectories. Finally, we demonstrate through extensive experiments that M-GRPO achieves superior stability and state-of-the-art performance on multiple reasoning benchmarks.

\section{Methodology}
\label{sec:method}
This chapter is structured to provide a clear and comprehensive overview of our proposed methodology. We begin in Section~\ref{sec:momentum} with a detailed description of the Momentum-anchored self-supervised reinforcement learning model. The subsequent section, Section~\ref{sec:iqr}, is dedicated to explaining the interquartile range (IQR) based method for trajectory entropy filtering. Finally, the complete, integrated algorithm is presented in Section~\ref{sec:code}.
\subsection{Momentum-anchored self-supervised RL}
\label{sec:momentum}
\begin{figure}[h]
    \centering
    \includegraphics[width=0.8\textwidth]{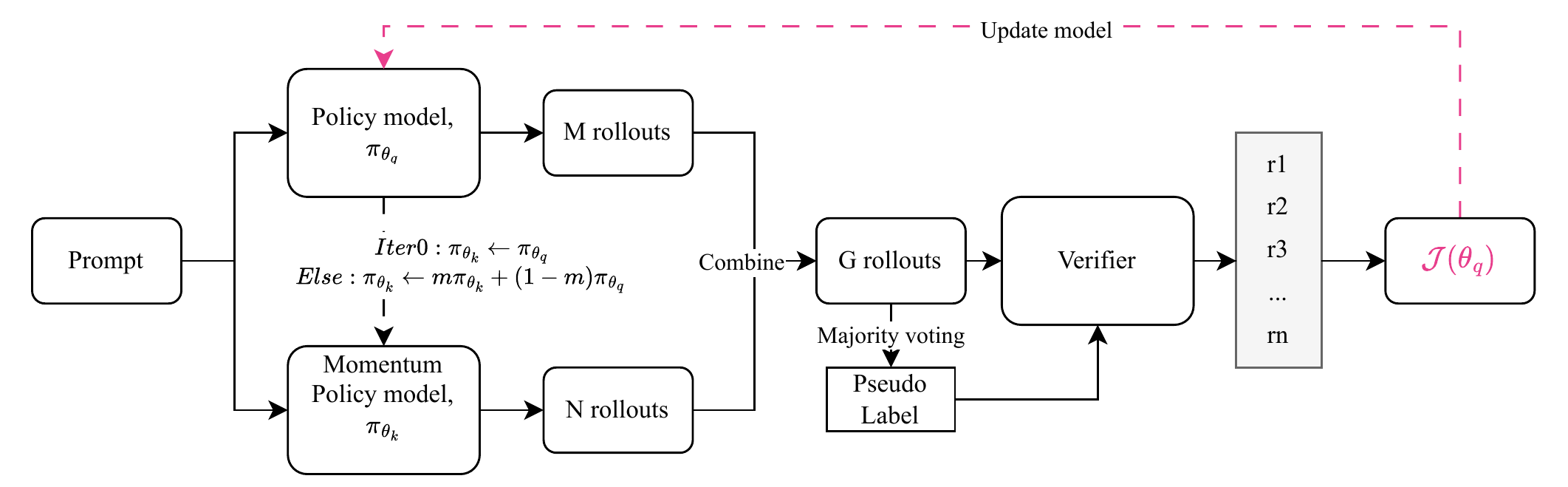}
    \caption{An illustration of our proposed momentum-based policy refinement framework. The  policy $\pi_{\theta_q}$ is trained using feedback derived from rollouts generated by both itself and a slowly evolving momentum policy $\pi_{\theta_k}$. The momentum's parameters are a moving average of the current policy model's, providing a stable target for pseudo-label generation via majority voting.}
    \label{fig:pipeline}
\vspace{-1em}
\end{figure}

To enhance the stability  of the self-supervised  process in policy refinement, we introduce a momentum-based RLVR framework, M-GRPO. The main framework is illustrated in Fig. ~\ref{fig:pipeline} Our approach is built upon the foundational principles of Group Relative Policy Optimization (GRPO)~\cite{shao2024deepseekmathpushinglimitsmathematical, yu2025dapoopensourcellmreinforcement}, but it incorporates a momentum-updated  model to provide a more consistent and reliable training signal. This method  stabilized through the momentum mechanism. The core idea is inspired by the success of momentum contrast in self-supervised visual representation learning ~\cite{chen2020improvedbaselinesmomentumcontrastive,he2020momentumcontrastunsupervisedvisual}, which we adapt to the context of policy optimization for self-supervised RLVR.

Formally, our framework consists of two models: the current policy model $\pi_{\theta_q}$ , which we aim to train, and a momentum-based  model $\pi_{\theta_k}$. The parameters $\theta_k$ of the momentum model are not updated by backpropagation. Instead, they are an exponential moving average of the current policy  model's parameters $\theta_q$. Following each training iteration of the current policy model, the momentum-model parameters are updated as follows:
\begin{equation}
    \pi_{\theta_k} \leftarrow m\pi_{\theta_k} + (1 - m)\pi_{\theta_q},
    \label{eqn:momentum_update}
\end{equation}
where $m \in [0, 1)$ is a momentum coefficient. A high value of $m$ (e.g., 0.99) ensures that the momentum model evolves slowly, thereby providing a stable source for generating reference outputs.

For each input prompt $x$ in a given training batch, we generate two distinct sets of rollouts. The current policy model policy $\pi_{\theta_q}(Y|x)$ samples $M$ responses, $\{y_i^q\}_{i=1}^M$, while the momentum policy model $\pi_{\theta_k}(Y|x)$ samples $N$ responses, $\{y_j^k\}_{j=1}^N$. These are combined to form a pool of $G = M + N$ total rollouts. 

From this combined pool, we identify a pseudo-ground truth response, $y_v$, using a majority voting mechanism. This mechanism selects the response that has the highest agreement with all other responses in the pool:
$y_v \leftarrow \arg\max_{y^* \in Y_{\text{pool}}} \sum_{y' \in Y_{\text{pool}}} \mathbb{I}[\text{ans}(y') = \text{ans}(y^*)]$, 
where $\mathbb{I}[\cdot]$ is the indicator function and $\text{ans}(\cdot)$ extracts the final answer from a response. The inclusion of the stable momentum model's rollouts in the voting pool is crucial for mitigating the noise and instability of pseudo-labels generated purely from the rapidly changing current policy model .With the pseudo-ground truth $y_v$ established, we can calculate the  reward   scores  of the current policy model's $M$ rollouts based on pseudo-ground truth $y_v$. The reward is binary: 1 for agreement and 0 for disagreement. Following the GRPO framework, we then compute the normalized advantage for each of the current policy model's sampled responses:
\begin{equation}
\hat{A}_{i} = \frac{r(y_v, y_i^q) - \text{mean}(\{r(y_v, y_j^q)\}_{j=1}^{M})}{\text{std}(\{r(y_v, y_j^q)\}_{j=1}^{M})}.
    \label{eqn:advantage}
\end{equation}
This advantage normalization is performed on a per-prompt basis over the current policy model's $M$ rollouts and serves to reduce the variance of the policy gradient estimate.

The final learning objective for the current policy model $\pi_{\theta_q}$ is to maximize the expected advantage-weighted log-likelihood of its generated responses:
\vspace{-1em}
\begin{equation}
    \mathcal{J}(\theta_q) = \mathbb{E}_{x\sim D, \{y_i^q\}_{i=1}^M \sim \pi_{\theta_q}(Y|x)} \left[ \sum_{i=1}^M \hat{A}_i \log \pi_{\theta_q}(y_i^q|x) \right].
    \label{eqn:objective}
\end{equation}

\subsection{IQR-based trajectory entropy filter}
\label{sec:iqr}
A critical challenge in self-supervised reinforcement learning is the management of trajectory quality. Trajectories with low entropy often correspond to policies that have prematurely converged or are overly confident, which can degrade the quality of the pseudo-labels used for training. To address this, we introduce a dynamic filtering mechanism based on the interquartile range (IQR) to prune low-entropy trajectories.

For each input sample $x$ from a mini-batch $\mathcal{B}$, we generate a total of $G$ rollouts, comprising $M$ trajectories from the current policy $\pi_{\theta_q}$ and $N$ trajectories from the momentum policy $\pi_{\theta_k}$. For each of these $G$ trajectories, we compute its trajectory-level entropy. The set of all trajectory entropies for a given input sample forms a distribution.

Rather than employing a static threshold \cite{zhang2025edgegrpoentropydrivengrpoguided}, such as removing the bottom 10\% of trajectories, which may be ill-suited for the dynamic nature of training, we utilize the IQR to identify and discard outliers. Specifically, for the set of $G$ entropy values associated with an input sample, we calculate the first quartile ($Q_1$) and the third quartile ($Q_3$). The interquartile range is then defined as $\text{IQR} = Q_3 - Q_1$. A trajectory is identified as a low-entropy outlier if its entropy falls below the threshold $Q_1 - k \cdot \text{IQR}$, where $k$ is a hyperparameter that controls the sensitivity of the filter. In our experiments, we set $k=0.75$.

This dynamic approach is particularly advantageous because the distribution of trajectory entropies can vary significantly, both across different input samples and throughout the training process. For instance, at the beginning of training, most trajectories may exhibit high entropy as the policy explores the action space. A static filter would be ineffective in this scenario, whereas our IQR-based method adapts to the local distribution of entropies, ensuring that only genuine outliers are removed. This leads to a more robust and stable training process by preserving high-quality, high-entropy trajectories for the subsequent learning steps.
\subsection{Pseudo-code}
\label{sec:code}
The overall training procedure for our Momentum-anchored GRPO (M-GRPO) with IQR-based filtering is summarized in Algorithm~\ref{alg:ours}. By leveraging a momentum policy model to stabilize the generation of pseudo-labels and a dynamic filtering mechanism to ensure their quality, our method provides a more consistent learning signal. This empirically leads to improved performance and training stability, effectively reducing the risk of policy degradation or collapse.
\label{sec:code}
\begin{algorithm}[H]
   \caption{M-GRPO with IQR-based Trajectory Entropy Filtering}
   \label{alg:ours}
   \begin{algorithmic}[1]
   \STATE {\bf Input:}  policy model $\pi_{\theta_q}$, training dataset $\mathcal{D}$, learning rate $\eta$, momentum coefficient $m$, number of policy model rollouts $M$, number of momentum rollouts $N$, total number of rollouts $G=M+N$, IQR coefficient $k=0.75$.
   \STATE Initialize momentum model parameters $\pi_{\theta_k} \leftarrow \pi_{\theta_q}$.
   \FOR{each training iteration}
       \STATE Sample a mini-batch of prompts $\mathcal{B} \subseteq \mathcal{D}$.
       \FOR{each prompt $x \in \mathcal{B}$}
           \STATE Generate $M$ responses from the current policy: $\{y_i^q\}_{i=1}^M \sim \pi_{\theta_q}(\cdot|x)$.
           \STATE Generate $N$ responses from the momentum model: $\{y_j^k\}_{j=1}^N \sim \pi_{\theta_k}(\cdot|x)$.
           \STATE Combine all generated responses: $\mathcal{Y} = \{y_i^q\}_{i=1}^M \cup \{y_j^k\}_{j=1}^N$.
           \STATE Calculate trajectory-level entropy for each response in $\mathcal{Y}$.
           \STATE Compute $Q_1$ and $Q_3$ for the set of trajectory entropies.
           \STATE Define the filtering threshold: $T_{IQR} = Q_1 - k \cdot (Q_3 - Q_1)$.
           \STATE Filter the set of responses: $\mathcal{Y}_{filtered} = \{y \in \mathcal{Y} \mid \text{entropy}(y) \geq T_{IQR}\}$.
           \STATE Identify the majority-voted pseudo-ground truth $y_v$ from $\mathcal{Y}_{filtered}$.
           \STATE Estimate the relative advantages $\hat{A}_i$ for responses from the current policy model in $\mathcal{Y}_{filtered}$ using Eq.~(\ref{eqn:advantage}).
       \ENDFOR
       \STATE Calculate the policy objective $\mathcal{J}(\theta_q)$ over the mini-batch using the filtered, high-quality trajectories and Eq.~(\ref{eqn:objective}).
       \STATE Update current policy parameters: $\theta_q \leftarrow \theta_q + \eta \nabla_{\theta_q} \mathcal{J}(\theta_q)$.
       \STATE Update momentum model parameters: $    \pi_{\theta_k} \leftarrow m\pi_{\theta_k} + (1 - m)\pi_{\theta_q}$ using Eq.~(\ref{eqn:momentum_update}).
   \ENDFOR
   \end{algorithmic}
\end{algorithm}
\section{Experiment}\label{sec:setting}
\textbf{Backbone Models and Baselines}. We utilize   Qwen3-4B-Base \cite{yang2025qwen3technicalreport} as  LLM backbone. We compare our result with  SRT~\cite{shafayat2025largereasoningmodelsselftrain}.

\textbf{Implementation Details}
We implement our algorithm on top of the VeRL framework ~\cite{sheng2024hybridflow}. Experiments are conducted on 8 × NVIDIA H200 GPUs. Specifically,  methods
are trained on the training split of the MATH dataset without its provided ground truth ~\cite{hendrycks2021measuringmathematicalproblemsolving}. We test all models' accuracy on the test split of MATH dataset as well.  For every RL update during training, we  generate 32
candidate rollouts per problem  and set temperature to 1.1 as default. We test model performance on MATH500~\cite{hendrycks2021measuringmathematicalproblemsolving}, AIME25 ~\cite{aime24,aime24_2}, GPQA Diamond ~\cite{gpqa_diamond,gpqa_diamond_2}, GPQA~\cite{gpqa_main},   Livecode~\cite{jain2024livecodebenchholisticcontaminationfree}, and  mbpp~\cite{austin2021programsynthesislargelanguage}. For AIME problems and GPQA Diamond problems, we sample 8 times and take the average accuracy. For the rest of therest of the benchmarks, we sample once.  For M-GRPO, momentum policy model rollouts N=$\frac{G}{4}$. For additional experimental details, please refer to Table~\ref{tab:exp_detail}. 
\section{Result}
\begin{table}[!t]
\centering
\resizebox{\textwidth}{!}{
\begin{tabular}{l|ccccccc}
\toprule[1.6pt]
\multirow{2}{*}{\textbf{}} & \multicolumn{3}{c}{\textbf{Math}} & \multicolumn{2}{c}{\textbf{Reasoning}} & \multicolumn{1}{c}{\textbf{Code}} \\
\cmidrule{2-7}
~ & \textbf{MATH500 } & \textbf{AIME24 } & \textbf{AIME25 } & \textbf{GPQA Dia} & \textbf{GPQA} & \textbf{LiveCode} \\
\midrule
\multicolumn{7}{c}{\textit{\textbf{Qwen3-4B-base}}} \\
\midrule
Original & 61.50\% & 0.83\% & 5.00\% & 34.41\% & 29.91\% & 9.61\% \\
\midrule
\textit{SRT\_Best} & 79.20\% & 12.50\% & 11.67\% & 38.26\% & 35.04\% & 19.69\% \\
\textit{SRT\_Final}  &  47.50\% & 7.50\% & 8.75\% & 28.54\% & 25.89\% & 16.12\% \\

\textit{SRT\_Final}  & 79.70\% & 12.50\% & 13.33\% & 37.94\% & \cellcolor{lightblue2}{37.72\%} & \cellcolor{lightblue2}{27.12\%} \\
\textit{M-GRPO+IQR\_Final} & \cellcolor{lightblue2}{79.75\%} & \cellcolor{lightblue2}{14.58\%} & \cellcolor{lightblue2}{14.17\%} & \cellcolor{lightblue2}{39.65\%} & 35.49\% & \textbackslash
\\
\bottomrule[1.6pt]
\end{tabular}
}
\caption{Main Results of RL performance comparison on reasoning benchmarks. Cell background colors indicate relative performance: darker colors denote better results within each model group. We compare M-GRPO with SRT~\cite{shafayat2025largereasoningmodelsselftrain} on Qwen3-4B-Base model trained on MATH~\cite{hendrycks2021measuringmathematicalproblemsolving} dataset without ground-truth label. \textit{SRT\_Best}  means we manually chosen the best checkpoint based on accuracy reward.  \textit{SRT\_Final}  means the final step checkpoint. M-GRPO show superior performance compared to SRT according to best accuracy and stability.}
\label{tab:rl_main} 
\vspace{-1em}
\end{table}
\vspace{-1em}
Our experimental results validate the effectiveness of M-GRPO in stabilizing the self-supervised reinforcement learning process and achieving superior performance compared to the baseline SRT method. 

\subsection{M-GRPO Prevents Policy Collapse and Achieves State-of-the-Art Performance}

A primary contribution of our work is the mitigation of "policy collapse," a critical failure mode in self-supervised RLVR. As illustrated in Fig~\ref{fig:srt-scaling-collapse} and further detailed in Fig~\ref{fig:srt-collapse}, the SRT~\cite{shafayat2025largereasoningmodelsselftrain} baseline exhibits a characteristic pattern of instability.  This instability necessitates manually selecting the best checkpoint (\texttt{SRT-Best}) before the performance crash, which is impractical for continuous training and deployment.

\begin{figure}[h]
    \centering    
    \includegraphics[width=1\textwidth]{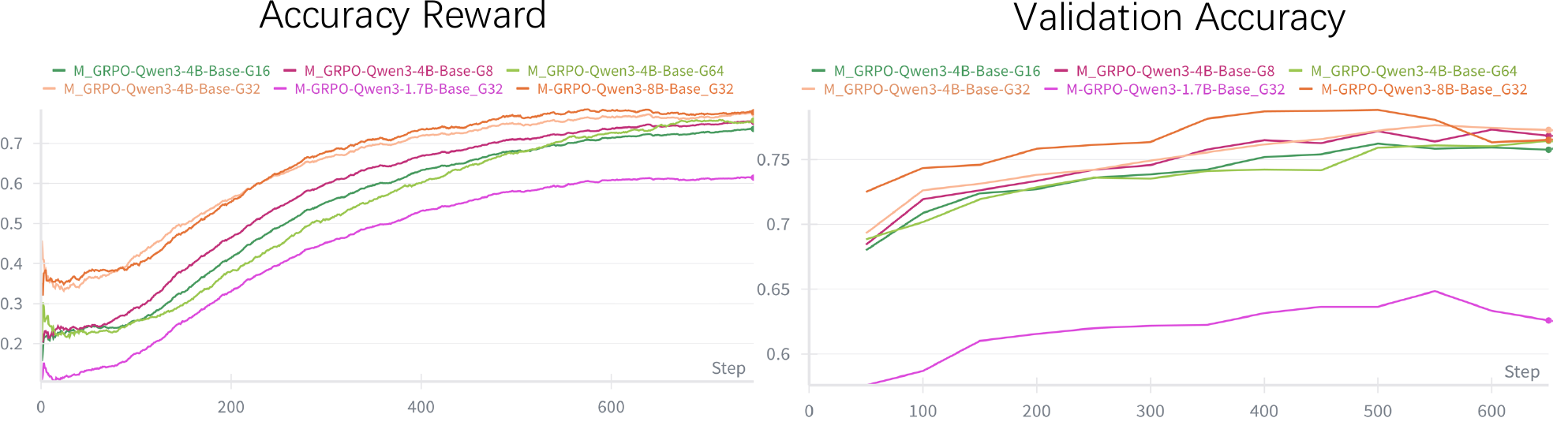}
    \caption{Training reward   on  the training split of \textbf{MATH} dataset and validation accuracy on test split  with different training backbone including Qwen3-4B-Base,Qwen3-1.7B-Base and Qwen3-8B-Base. 
    Our proposed M-GRPO sustains stable reward signals throughout training. }
    \label{fig:result}
\end{figure}

In stark contrast, M-GRPO demonstrates remarkable training stability. As shown in Fig~\ref{fig:result}, M-GRPO sustains an improving stable reward throughout the training horizon, which translates directly to a consistently high validation accuracy without any signs of degradation. This stability obviates the need for manual intervention or checkpoint cherry-picking.

\begin{table}[!t]
\centering
\resizebox{\textwidth}{!}{
\begin{tabular}{l|cccccccc}
\toprule[1.6pt]
\multirow{2}{*}{\textbf{}} & \multicolumn{3}{c}{\textbf{Math}} & \multicolumn{3}{c}{\textbf{Reasoning}} & \multicolumn{1}{c}{\textbf{Code}} \\
\cmidrule{2-8}
~ & \textbf{MATH500 } & \textbf{AIME24 } & \textbf{AIME25 } & \textbf{GPQA Dia} & \textbf{GPQA} & \textbf{MMLU-pro}&\textbf{mbpp} \\
\midrule
\multicolumn{7}{c}{\textit{\textbf{Qwen3-4B-base}}} \\
\midrule
Original & 61.50\% & 0.83\% & 5.00\% & 34.41\% &  29.91\% &51.38\% & 63.40\% \\
\midrule
 \textit{M-GRPO+IQR\_{G8}}& 77.60\% & 11.25\% & 10.42\% & 39.02\% & 37.50\% & 56.05\% &68.60\%\\
 \textit{M-GRPO+IQR\_{G16}} & 79.75\% & 14.43\% & 10.00\% & 39.65\% & 33.94\% & 57.05\% &70.40\%\\
 \textit{M-GRPO+IQR\_{G32}} & 79.75\% & 14.58\% & 14.17\% & 39.65\% & 35.49\% & 55.47\% &70.60\% \\
 \textit{M-GRPO+IQR\_{G256}} & 79.50\% & 16.67\% & 14.17\% & 40.66\% & 38.39\% & 55.08\% &70.40\%
\\
\bottomrule[1.6pt]
\end{tabular}
}
\caption{Scaling analysis of the M-GRPO+IQR method on math, reasoning, and code generation benchmarks. This table illustrates the effect of increasing the number of rollouts ($G$) on model performance. A clear trend of improvement is observed as $G$ is scaled from 8 to 32. However, the performance gains begin to plateau beyond this point, with only minimal improvements observed when increasing $G$ to 256.}
\label{tab:rl_main_scaling} 
\vspace{-1em}
\end{table}
The quantitative results of this stability are presented in Table~\ref{tab:rl_main}. At the final training checkpoint, \texttt{M-GRPO-Final} significantly outperforms \texttt{SRT-Final} across all six reasoning benchmarks, highlighting the severe performance loss SRT suffers from collapse. More importantly, \texttt{M-GRPO-Final} is highly competitive with, and frequently surpasses, the manually selected \texttt{SRT-Best}. For instance, on GPQA, M-GRPO achieves a final accuracy of 40.09\%, compared to SRT's peak of 35.04\%. The improvements are observed on more challenging benchmarks such as AIME24 (+2.92\%), GPQA (+5.05\%), and LiveCode (+7.43\%) in absolute accuracy. These results confirm that M-GRPO not only stabilizes training but also enables the model to converge to a more robust and capable state.

\subsection{Analysis of Rollout Scaling}

To further understand the dynamics of self-supervised RLVR, we analyzed the impact of varying the number of rollouts (G) on the SRT baseline's performance and stability. As shown in Table~\ref{tab:ssl_scaling_srt}, increasing the number of rollouts can indeed improve the peak performance of SRT. For example, on MATH500, scaling from G=16 to G=64 boosts the best-achieved accuracy from 76.15\% to 79.20\%. For M-GRPO, as illustrated in Table~\ref{tab:rl_main_scaling}, a clear trend of improvement is observed as $G$ is scaled from 8 to 32. However, the performance gains begin to plateau beyond this point, with only minimal improvements observed when increasing $G$ to 256. 

\section{Conclusion}
In this paper, we have identified a critical instability issue in self-supervised reinforcement learning for large language models, leading to policy collapse during extended training. We have demonstrated that this failure is attributable to the inherent lack of a stable target in self-rewarding systems. To address this, we introduced M-GRPO, a momentum-anchored, majority-voting framework that stabilizes the learning process.

{
    \small
    \bibliographystyle{plainnat}
    \bibliography{neurips_2025}

@misc{shafayat2025largereasoningmodelsselftrain,
      title={Can Large Reasoning Models Self-Train?}, 
      author={Sheikh Shafayat and Fahim Tajwar and Ruslan Salakhutdinov and Jeff Schneider and Andrea Zanette},
      year={2025},
      eprint={2505.21444},
      archivePrefix={arXiv},
      primaryClass={cs.LG},
      url={https://arxiv.org/abs/2505.21444}, 
}

@misc{tan2025scalingbehaviorsllmreinforcement,
      title={Scaling Behaviors of LLM Reinforcement Learning Post-Training: An Empirical Study in Mathematical Reasoning}, 
      author={Zelin Tan and Hejia Geng and Mulei Zhang and Xiaohang Yu and Guancheng Wan and Yifan Zhou and Qiang He and Xiangyuan Xue and Heng Zhou and Yutao Fan and Zhongzhi Li and Zaibin Zhang and Guibin Zhang and Chen Zhang and Zhenfei Yin and Lei Bai},
      year={2025},
      eprint={2509.25300},
      archivePrefix={arXiv},
      primaryClass={cs.LG},
      url={https://arxiv.org/abs/2509.25300}, 
}

@misc{yang2025qwen3technicalreport,
      title={Qwen3 Technical Report}, 
      author={An Yang and Anfeng Li and Baosong Yang and Beichen Zhang and Binyuan Hui and Bo Zheng and Bowen Yu and Chang Gao and Chengen Huang and Chenxu Lv and Chujie Zheng and Dayiheng Liu and Fan Zhou and Fei Huang and Feng Hu and Hao Ge and Haoran Wei and Huan Lin and Jialong Tang and Jian Yang and Jianhong Tu and Jianwei Zhang and Jianxin Yang and Jiaxi Yang and Jing Zhou and Jingren Zhou and Junyang Lin and Kai Dang and Keqin Bao and Kexin Yang and Le Yu and Lianghao Deng and Mei Li and Mingfeng Xue and Mingze Li and Pei Zhang and Peng Wang and Qin Zhu and Rui Men and Ruize Gao and Shixuan Liu and Shuang Luo and Tianhao Li and Tianyi Tang and Wenbiao Yin and Xingzhang Ren and Xinyu Wang and Xinyu Zhang and Xuancheng Ren and Yang Fan and Yang Su and Yichang Zhang and Yinger Zhang and Yu Wan and Yuqiong Liu and Zekun Wang and Zeyu Cui and Zhenru Zhang and Zhipeng Zhou and Zihan Qiu},
      year={2025},
      eprint={2505.09388},
      archivePrefix={arXiv},
      primaryClass={cs.CL},
      url={https://arxiv.org/abs/2505.09388}, 
}

@article{sheng2024hybridflow,
  title   = {HybridFlow: A Flexible and Efficient RLHF Framework},
  author  = {Guangming Sheng and Chi Zhang and Zilingfeng Ye and Xibin Wu and Wang Zhang and Ru Zhang and Yanghua Peng and Haibin Lin and Chuan Wu},
  year    = {2024},
  journal = {arXiv preprint arXiv: 2409.19256}
}

@misc{austin2021programsynthesislargelanguage,
      title={Program Synthesis with Large Language Models}, 
      author={Jacob Austin and Augustus Odena and Maxwell Nye and Maarten Bosma and Henryk Michalewski and David Dohan and Ellen Jiang and Carrie Cai and Michael Terry and Quoc Le and Charles Sutton},
      year={2021},
      eprint={2108.07732},
      archivePrefix={arXiv},
      primaryClass={cs.PL},
      url={https://arxiv.org/abs/2108.07732}, 
}

@misc{zhang2025edgegrpoentropydrivengrpoguided,
      title={EDGE-GRPO: Entropy-Driven GRPO with Guided Error Correction for Advantage Diversity}, 
      author={Xingjian Zhang and Siwei Wen and Wenjun Wu and Lei Huang},
      year={2025},
      eprint={2507.21848},
      archivePrefix={arXiv},
      primaryClass={cs.AI},
      url={https://arxiv.org/abs/2507.21848}, 
}

@misc{cui2025entropymechanismreinforcementlearning,
      title={The Entropy Mechanism of Reinforcement Learning for Reasoning Language Models}, 
      author={Ganqu Cui and Yuchen Zhang and Jiacheng Chen and Lifan Yuan and Zhi Wang and Yuxin Zuo and Haozhan Li and Yuchen Fan and Huayu Chen and Weize Chen and Zhiyuan Liu and Hao Peng and Lei Bai and Wanli Ouyang and Yu Cheng and Bowen Zhou and Ning Ding},
      year={2025},
      eprint={2505.22617},
      archivePrefix={arXiv},
      primaryClass={cs.LG},
      url={https://arxiv.org/abs/2505.22617}, 
}

@misc{hu2025brorlscalingreinforcementlearning,
      title={BroRL: Scaling Reinforcement Learning via Broadened Exploration}, 
      author={Jian Hu and Mingjie Liu and Ximing Lu and Fang Wu and Zaid Harchaoui and Shizhe Diao and Yejin Choi and Pavlo Molchanov and Jun Yang and Jan Kautz and Yi Dong},
      year={2025},
      eprint={2510.01180},
      archivePrefix={arXiv},
      primaryClass={cs.LG},
      url={https://arxiv.org/abs/2510.01180}, 
}

@misc{jain2024livecodebenchholisticcontaminationfree,
      title={LiveCodeBench: Holistic and Contamination Free Evaluation of Large Language Models for Code}, 
      author={Naman Jain and King Han and Alex Gu and Wen-Ding Li and Fanjia Yan and Tianjun Zhang and Sida Wang and Armando Solar-Lezama and Koushik Sen and Ion Stoica},
      year={2024},
      eprint={2403.07974},
      archivePrefix={arXiv},
      primaryClass={cs.SE},
      url={https://arxiv.org/abs/2403.07974}, 
}

@misc{hendrycks2021measuringmathematicalproblemsolving,
      title={Measuring Mathematical Problem Solving With the MATH Dataset}, 
      author={Dan Hendrycks and Collin Burns and Saurav Kadavath and Akul Arora and Steven Basart and Eric Tang and Dawn Song and Jacob Steinhardt},
      year={2021},
      eprint={2103.03874},
      archivePrefix={arXiv},
      primaryClass={cs.LG},
      url={https://arxiv.org/abs/2103.03874}, 
}

@misc{aime24,
      title={AIME. AIME problems and solutions,}, 
      year={2025},
      url={https://artofproblemsolving.com/wiki/index.php/AIME_Problems_and_Solutions}, 
}

@misc{gpqa_diamond,
      title={GPQA Diamond}, 
      year={2025},
      url={https://epoch.ai/benchmarks/gpqa-diamond}
}

@misc{gpqa_diamond_2,
      title={GPQA Diamond}, 
      year={2025},
      url={https://huggingface.co/datasets/fingertap/GPQA-Diamond }
}

@misc{gpqa_main,
      title={GPQA: A Graduate-Level Google-Proof Q&A Benchmark}, 
      author={David Rein and Betty Li Hou and Asa Cooper Stickland and Jackson Petty and Richard Yuanzhe Pang and Julien Dirani and Julian Michael and Samuel R. Bowman},
      year={2023},
      eprint={2311.12022},
      archivePrefix={arXiv},
      primaryClass={cs.AI},
      url={https://arxiv.org/abs/2311.12022}, 
}

@misc{aime24_2,
      title={AIME. AIME problems and solutions,}, 
      year={2025},
      url={https://huggingface.co/datasets/Maxwell-Jia/AIME_2024}, 
}

@misc{chen2020improvedbaselinesmomentumcontrastive,
      title={Improved Baselines with Momentum Contrastive Learning}, 
      author={Xinlei Chen and Haoqi Fan and Ross Girshick and Kaiming He},
      year={2020},
      eprint={2003.04297},
      archivePrefix={arXiv},
      primaryClass={cs.CV},
      url={https://arxiv.org/abs/2003.04297}, 
}

@misc{he2020momentumcontrastunsupervisedvisual,
      title={Momentum Contrast for Unsupervised Visual Representation Learning}, 
      author={Kaiming He and Haoqi Fan and Yuxin Wu and Saining Xie and Ross Girshick},
      year={2020},
      eprint={1911.05722},
      archivePrefix={arXiv},
      primaryClass={cs.CV},
      url={https://arxiv.org/abs/1911.05722}, 
}

@article{Zuo2025TTRLTR,
  title={TTRL: Test-Time Reinforcement Learning},
  author={Yuxin Zuo and Kaiyan Zhang and Shang Qu and Li Sheng and Xuekai Zhu and Biqing Qi and Youbang Sun and Ganqu Cui and Ning Ding and Bowen Zhou},
  journal={ArXiv},
  year={2025},
  volume={abs/2504.16084},
  url={https://api.semanticscholar.org/CorpusID:277993666}
}

@misc{yu2025dapoopensourcellmreinforcement,
      title={DAPO: An Open-Source LLM Reinforcement Learning System at Scale}, 
      author={Qiying Yu and Zheng Zhang and Ruofei Zhu and Yufeng Yuan and Xiaochen Zuo and Yu Yue and Weinan Dai and Tiantian Fan and Gaohong Liu and Lingjun Liu and Xin Liu and Haibin Lin and Zhiqi Lin and Bole Ma and Guangming Sheng and Yuxuan Tong and Chi Zhang and Mofan Zhang and Wang Zhang and Hang Zhu and Jinhua Zhu and Jiaze Chen and Jiangjie Chen and Chengyi Wang and Hongli Yu and Yuxuan Song and Xiangpeng Wei and Hao Zhou and Jingjing Liu and Wei-Ying Ma and Ya-Qin Zhang and Lin Yan and Mu Qiao and Yonghui Wu and Mingxuan Wang},
      year={2025},
      eprint={2503.14476},
      archivePrefix={arXiv},
      primaryClass={cs.LG},
      url={https://arxiv.org/abs/2503.14476}, 
}

@misc{zhang2025corewardselfsupervisedreinforcementlearning,
      title={Co-Reward: Self-supervised Reinforcement Learning for Large Language Model Reasoning via Contrastive Agreement}, 
      author={Zizhuo Zhang and Jianing Zhu and Xinmu Ge and Zihua Zhao and Zhanke Zhou and Xuan Li and Xiao Feng and Jiangchao Yao and Bo Han},
      year={2025},
      eprint={2508.00410},
      archivePrefix={arXiv},
      primaryClass={cs.LG},
      url={https://arxiv.org/abs/2508.00410}, 
}

@misc{shao2024deepseekmathpushinglimitsmathematical,
      title={DeepSeekMath: Pushing the Limits of Mathematical Reasoning in Open Language Models}, 
      author={Zhihong Shao and Peiyi Wang and Qihao Zhu and Runxin Xu and Junxiao Song and Xiao Bi and Haowei Zhang and Mingchuan Zhang and Y. K. Li and Y. Wu and Daya Guo},
      year={2024},
      eprint={2402.03300},
      archivePrefix={arXiv},
      primaryClass={cs.CL},
      url={https://arxiv.org/abs/2402.03300}, 
}

@misc{zhao2025learningreasonexternalrewards,
      title={Learning to Reason without External Rewards}, 
      author={Xuandong Zhao and Zhewei Kang and Aosong Feng and Sergey Levine and Dawn Song},
      year={2025},
      eprint={2505.19590},
      archivePrefix={arXiv},
      primaryClass={cs.LG},
      url={https://arxiv.org/abs/2505.19590}, 
}
}
\clearpage
\appendix
\section{Appendix}

  \begin{figure}[h]
    \centering    \includegraphics[width=1\textwidth]{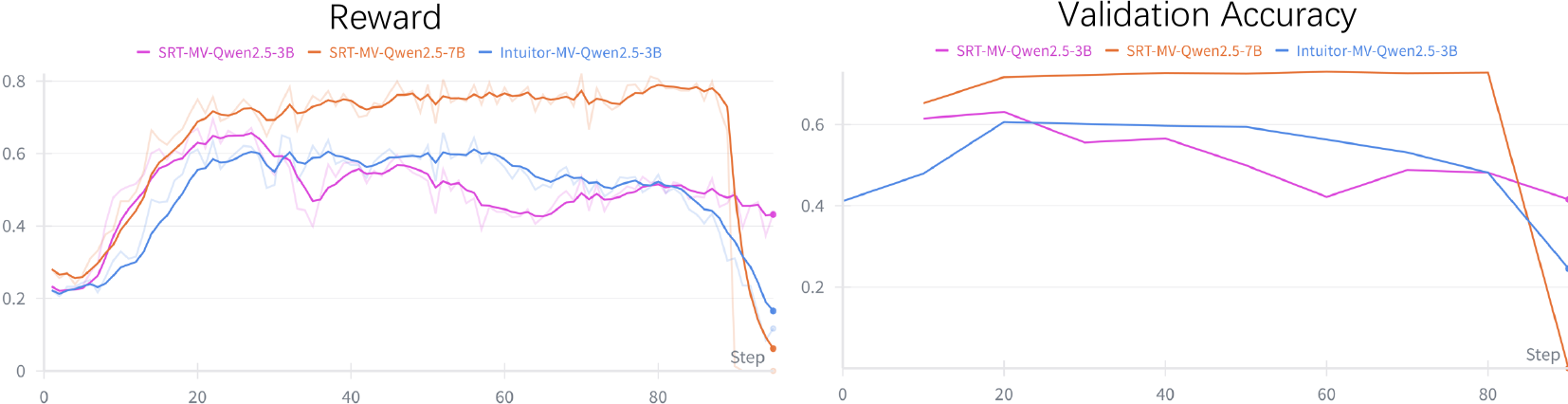}
    \caption{Reproduction  of \textcolor{magenta}{SRT-MV-Qwen2.5-3B}, 
    \textcolor{orange}{SRT-MV-Qwen2.5-7B} based on SRT~\cite{shafayat2025largereasoningmodelsselftrain}, 
    and \textcolor{blue}{Intuitor-MV-Qwen2.5-3B} based on Intuitor ~\cite{zhao2025learningreasonexternalrewards}, which all self-supervised training on train split of  MATH dataset ~\cite{hendrycks2021measuringmathematicalproblemsolving} and validate on test split of MATH dataset. The training reward rises initially and then precipitously or progressively  crashes  , accompanied by degradation in validation accuracy on the test set.}
    \label{fig:srt-collapse}
\end{figure}  

\clearpage
\subsection{More  experiment details}
\begin{table*}[h]
\caption{More detailed experimental parameter settings.}
\begin{center}
\resizebox{0.9\columnwidth}{!}{
\begin{tabular}{ll}
\toprule[1.5pt]
\multicolumn{2}{l}{\textbf{Training   Configuration}}           \\
\midrule[0.6pt]
Train Batch Size (Number of Sampled Questions)         & 8      \\
Max Prompt Length        & 512      \\
Max Response Length      & 3072     \\
Clip Ratio               & 0.2      \\
\midrule[0.6pt]
\multicolumn{2}{l}{\textbf{Optimizer Parameters}}          \\
\midrule[0.6pt]
Optimizer                & AdamW ($\beta_1=0.9$, $\beta_2=0.999$, $\epsilon=10^{-8}$)                 \\
Learning Rate            & 1e-06 \\
Warmup Style             & Cosine   \\
Warmup Steps Ratio       & 0.1      \\
KL Loss Coefficient      & 0.005    \\
\midrule[0.6pt]
\multicolumn{2}{l}{\textbf{Temperature}}           \\
\midrule[0.6pt]
Training Temperature & 1.1 \\
Evaluation Temperature & 0.8 \\
\bottomrule[1.5pt]
\end{tabular}}
\label{tab:exp_detail}
\end{center}
\end{table*}

\end{document}